\documentclass[10pt,twocolumn,letterpaper]{article}

\usepackage{multirow}
\usepackage{color}
\usepackage{wacv}              

\usepackage{stfloats}

\definecolor{wacvblue}{rgb}{0.21,0.49,0.74}
\usepackage[pagebackref,breaklinks,colorlinks,allcolors=wacvblue]{hyperref}

\def\wacvPaperID{808} 
\def\confName{WACV}
\def\confYear{2027}

\title{Latent Spectral Prompting for Multimodal Hyperspectral Object Tracking}

\author{Wangshu Cai\textsuperscript{$1$}\qquad Jun Zhou\textsuperscript{$2$}\thanks{Corresponding authors.} \qquad Lei Wang\textsuperscript{$1*$} \\ Mohammad Aminul Islam\textsuperscript{$3$} \qquad Xu Han\textsuperscript{$1$} \qquad Kuldip Paliwal\textsuperscript{$1$}\\
$^1$Griffith University, Australia, $^2$East China Normal University, China, \\ $^3$Bangladesh Agricultural University, Bangladesh\\
\{wangshu.cai, xu.han2\}@griffithuni.edu.au, \{l.wang4, k.paliwal\}@griffith.edu.au, \\
amin@bau.edu.bd, jzhou@ee.ecnu.edu.cn
}

\makeatletter
\renewcommand{\maketitlesupplementary}{
    \newpage
    \twocolumn[
        \centering
        \Large
        \textbf{\thetitle}\\
        \vspace{0.5em}
        Supplementary Material\\
        \vspace{1em}

        {\large
        Wangshu Cai\textsuperscript{1}
        \qquad Jun Zhou\textsuperscript{2*}
        \qquad Lei Wang\textsuperscript{1*}\\
        Mohammad Aminul Islam\textsuperscript{3}
        \qquad Xu Han\textsuperscript{1}
        \qquad Kuldip Paliwal\textsuperscript{1}\\[0.5em]

        \textsuperscript{1}Griffith University, Australia,
        \textsuperscript{2}East China Normal University, China,\\
        \textsuperscript{3}Bangladesh Agricultural University, Bangladesh\\
        \{wangshu.cai, xu.han2\}@griffithuni.edu.au,
        \{l.wang4, k.paliwal\}@griffith.edu.au,\\
        amin@bau.edu.bd, jzhou@ee.ecnu.edu.cn\\
        \textsuperscript{*}Corresponding authors.
        }
        \vspace{1em}
    ]
}
\makeatother

\begin{document}
\maketitle

\begin{abstract}

Hyperspectral videos provide rich spectral signatures for robust tracking under challenging conditions such as camouflage and low illumination. However, limited hyperspectral annotations and the large RGB-hyperspectral domain gap hinder the direct use of modern temporally pretrained RGB trackers.  We propose \textit{SProTrack}, a multimodal framework that transfers temporal knowledge from pretrained RGB trackers to the hyperspectral domain through latent spectral prompting. The key idea is to decouple temporal modeling from spectral adaptation, allowing temporal dynamics to be inherited from RGB foundation models while hyperspectral information is learned from unlabeled data in a self-supervised manner. Unlike band-selection methods that discard most spectral information or prompt-based approaches that rely on local-search crops, SProTrack learns a global, scene-level spectral representation from the full hyperspectral frame using an autoencoder. This latent representation is used to guide spectral prompt extraction while a pretrained RGB tracker preserves strong temporal modeling on RGB-compatible inputs. By encoding spectral-spatial structure into compact latent prompts, SProTrack  bridges the RGB-hyperspectral modality gap without extensive retraining or dense supervision. 
Experiments on HOT2020 and HOT2023 demonstrate that SProTrack
achieves the best overall performance, with strong results across
all spectral domains.

\end{abstract}

\section{Introduction}
\label{sec:intro}

\begin{figure}[tbp]
    \centering
    \includegraphics[width=0.8\columnwidth]{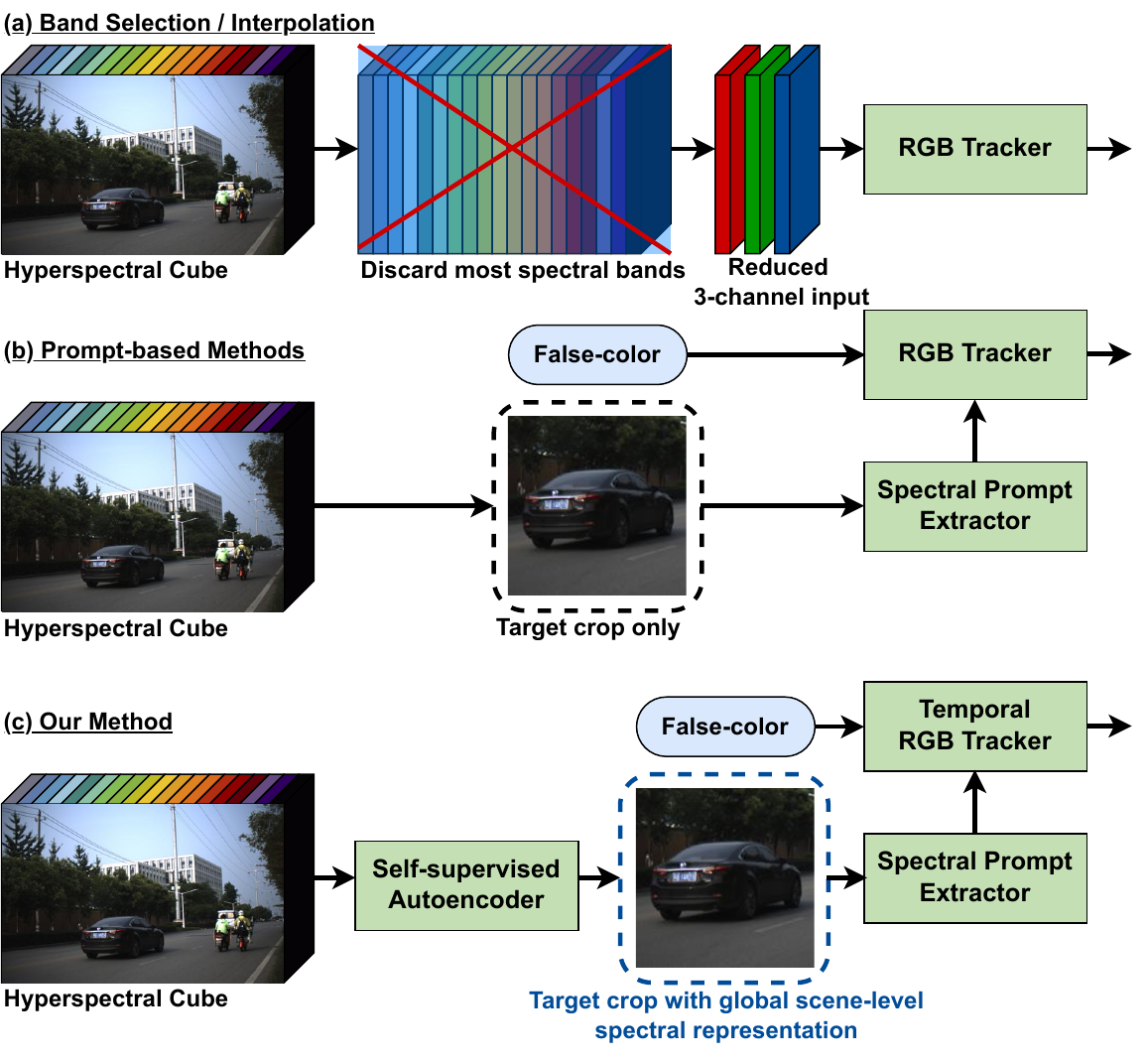}
    \vspace{-0.2cm}
    \caption{Hyperspectral tracking paradigms. (a) Band selection discards most bands, using a reduced 3-channel input. (b) Prompt-based methods keep all bands but extract spectral features only from local crops. (c) Ours encodes the full frame 
    into global scene-level spectral prompts for a temporal RGB tracker. }
    \label{fig:teaser}
\end{figure}

Visual object tracking is a fundamental problem in computer vision, aiming to localize a target object across video frames given its initial state. Recent advances, driven by large-scale RGB datasets and transformer-based architectures, have significantly improved tracking performance, particularly in modeling long-range temporal dynamics \cite{ostrack, chen2023seqtrack, lorat,raj2025tracknetv4,sharma2026rethinking}. However, RGB imagery is inherently limited to three spectral channels, making it vulnerable to challenging scenarios such as low illumination, camouflage, and appearance ambiguity. To compensate for this spectral limitation, Hyperspectral imaging offers a compelling alternative by capturing dense spectral information across dozens or hundreds of wavelength bands. These rich spectral signatures encode intrinsic material properties that are invariant to appearance changes, providing strong discriminative cues for robust tracking. Despite this advantage, hyperspectral object tracking (HOT) remains underexplored compared to its RGB counterpart \cite{landgrebe2002hyperspectral, plaza2009recent}. The primary bottleneck lies in the scarcity of annotated hyperspectral video data, which makes it difficult to train modern deep tracking models without severe overfitting.

A natural solution is to transfer knowledge from pretrained RGB trackers \cite{baenet, tang2022robustbaht}. However, this introduces a fundamental challenge: \emph{the spectral domain gap}. Existing approaches bridge this gap either by reducing hyperspectral data to three channels via band selection or interpolation \cite{sun2019hyperspectral, wang2018optimal, feng2017hyperspectral, ying2021unaligned}, or by directly learning spectral representations through fusion or prompting \cite{chen2024phtrack, wu2025mp,he2025hyperspectral,kong2024joint} and integrating them with pre-trained RGB trackers in an end-to-end manner. The overviews of the two methods are shown in \cref{fig:teaser} (a) and (b). Band selection methods inevitably discard valuable inter-band correlations, while end-to-end approaches jointly learn spectral representations and temporal dynamics from limited data, leading to suboptimal generalization. More recent prompt-based methods mitigate this issue by injecting auxiliary modality information into pretrained RGB backbones \cite{vipt, dasspnet, sun2024joint}. Nevertheless, these methods typically operate on local search regions and lack explicit temporal modeling, limiting their ability to fully exploit both spectral and temporal information \cite{spirit, wang2025hyperspectral}.

A common thread runs through these failures: band selection, end-to-end fusion, and local prompting all entangle spectral adaptation with temporal modeling, forcing both to be learned jointly from scarce data. We therefore revisit hyperspectral tracking from a multimodal perspective, where the key challenge is not merely adapting hyperspectral inputs but decoupling and integrating complementary information across modalities: inheriting strong temporal priors from RGB foundation models while learning robust spectral representations from limited hyperspectral data (see \cref{fig:teaser}(c)). Building on this insight, we propose SProTrack, a multimodal tracking framework that transfers temporal knowledge from pretrained RGB trackers through latent spectral prompting in two phases. First, a Spectral-Spatial AutoEncoder (SSAE) is trained in a self-supervised manner on full hyperspectral frames, exploiting abundant unlabeled data to learn compact latent representations that capture global scene context beyond the target box. Second, these representations are injected as visual prompts into a frozen, temporally-aware RGB backbone, inheriting its temporal dynamics while adapting to hyperspectral inputs through lightweight LoRA \cite{hu2022lora} modulation.

These three components address the limitations identified above in a unified manner. First, decoupling spectral representation learning from temporal modeling avoids the overfitting that arises when end-to-end methods learn both jointly from scarce annotations. Second, injecting latent spectral prompts enables deep, layer-wise multimodal fusion without modifying the backbone, in contrast to the shallow, input-level fusion of prior prompt-based methods. Third, full-frame self-supervised learning lets the model exploit the global scene context that band-selection and local-crop approaches discard.
Our main \textbf{contributions} are:
\renewcommand{\labelenumi}{\roman{enumi}.}

\begin{enumerate}[leftmargin=0.5cm]

\item We formulate hyperspectral object tracking as a \emph{multimodal adaptation problem} and propose a decoupled learning paradigm that separates temporal modeling from spectral representation learning, mitigating overfitting under limited annotations.

\item We introduce a \emph{Spectral-Spatial AutoEncoder (SSAE)} that learns compact latent representations from full hyperspectral frames in a self-supervised manner, capturing global scene-level context and preserving inter-band correlations beyond local regions.

\item We propose a \emph{latent spectral prompting} that injects layer-wise representations into a frozen, temporally-aware RGB backbone, achieving deep multimodal fusion while retaining strong temporal priors, and yielding strong performance across all spectral domains. 

\end{enumerate}

\section{Related Work}
\label{sec:relatedwork}

\noindent\textbf{Hyperspectral object tracking.} Early HOT methods are primarily based on generative models and handcrafted spectral features, such as spectral angle mapper (SAM) and radiative transfer modeling. These approaches explicitly model reflectance properties but often fail to capture complex spatial-temporal dynamics, leading to limited robustness in challenging scenarios \cite{chakravarty2021hyperspectral, liu2022siamhyper, xiong2020material}.
More recent works adopt discriminative learning frameworks. For instance, channel selection strategies reduce spectral redundancy by selecting informative bands, improving efficiency but discarding inter-band correlations \cite{zhang2023fast, yellin2024concurrent, li2023jointly}. Attention-based models further model spectral dependencies through hierarchical or hybrid attention, but still suffer from the trade-off between spectral richness and noise redundancy \cite{liu2021anchor, liu2024deep, liu2022siamhyper}.
Unlike these approaches that either compress spectral information heuristically or entangle spectral-spatial modeling within a single network, our method learns a \emph{compact latent spectral representation} from full hyperspectral frames in a self-supervised manner, preserving global spectral structure without manual band selection or heavy architectural coupling.

\noindent\textbf{Cross-modal transfer \& RGB-based tracking adaptation.} Given the scarcity of hyperspectral annotations, a dominant paradigm is to transfer knowledge from RGB trackers pretrained on large-scale datasets \cite{3dcae, dua2021convolution, ravi2024sam2segmentimages,karaev2024cotracker}. Early approaches adopt full fine-tuning, directly adapting RGB models to hyperspectral inputs, but often suffer from overfitting under insufficient data. To address this, recent works explore parameter-efficient adaptation, hybrid architectures that bridge spectral and spatial features, and domain adaptation modules that learn domain-invariant representations across sensors and environments \cite{sun2023siamohot, 10491347transdat}.
Existing transfer-based methods typically \emph{jointly learn spectral representation and temporal dynamics}, which leads to suboptimal learning under limited data. In contrast, we explicitly \emph{decouple temporal modeling from spectral adaptation}: a frozen RGB backbone preserves strong temporal priors, while spectral information is injected through lightweight latent prompts, improving data efficiency and avoiding catastrophic forgetting.

\noindent\textbf{Prompt learning \& multimodal representation.} Prompt learning has recently emerged as an effective paradigm for adapting large pretrained models with minimal parameters. In visual tracking, early works such as ProTrack~\cite{yang2022prompting} introduce prompt-based modulation, and subsequent methods extend this to hyperspectral tracking by learning modality-specific prompts while keeping the backbone frozen \cite{zhou2022learning, guo2024sptrack, hdsp, he2025hyperspectral, cai2023robust}. A parallel line explicitly models spectral-spatial-temporal interactions with transformer or state-space architectures \cite{wang2024trackingmamba, xiong2025suit, li2025unified}, but often requires training complex models from scratch, which limits scalability under data scarcity.
Despite their success, existing prompt-based methods exhibit three key limitations:
(i) \emph{Locality}: prompts are typically learned from local search regions, limiting global spectral context;
(ii) \emph{Shallow fusion}: prompts are often injected at the input level, lacking deep integration across layers;
(iii) \emph{Lack of explicit representation learning}: spectral information is not modeled in a structured latent space.
Our method addresses these limitations through a \emph{latent spectral prompting mechanism} learned via a self-supervised autoencoder on full frames, enabling global, compact representations injected into the backbone for deeper multimodal fusion.

\begin{figure*}[tbp]
    \centering
    \includegraphics[width=0.88\textwidth]{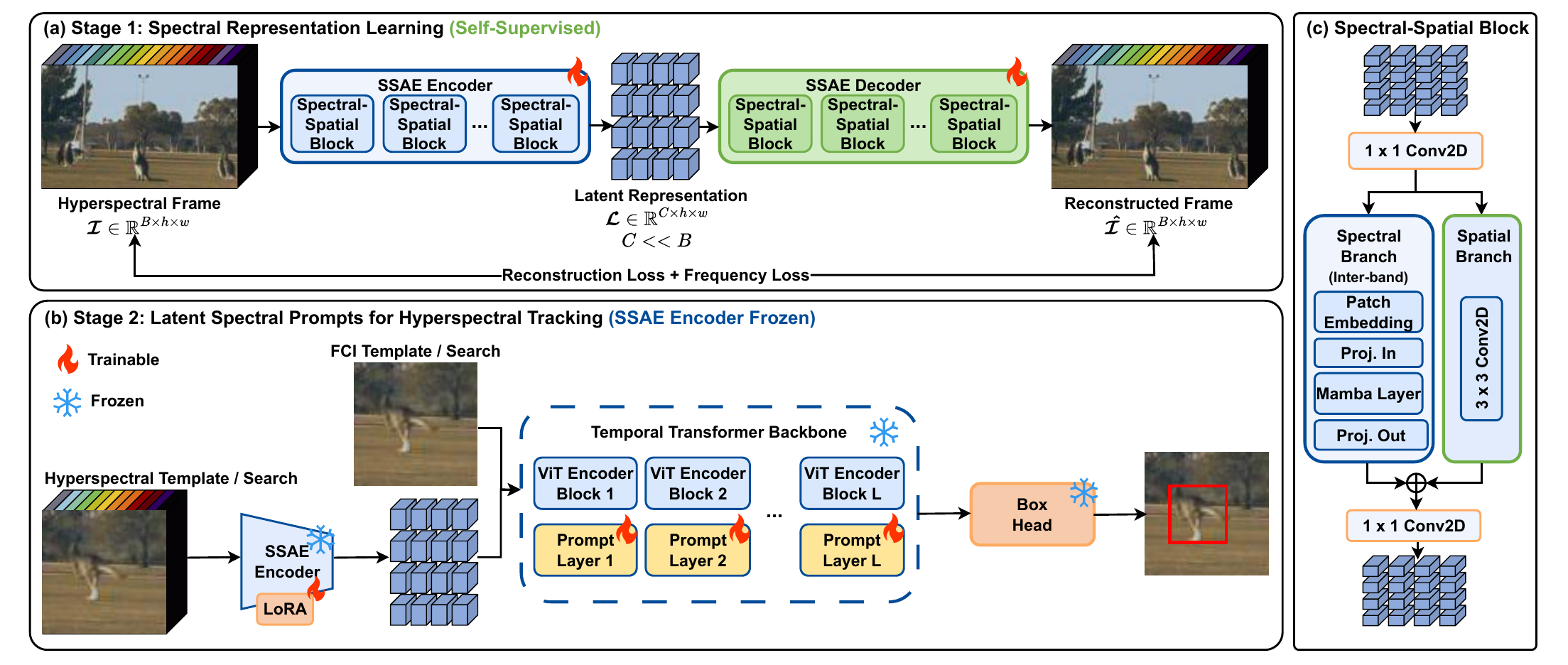}
    \vspace{-0.2cm}
    \caption{Overview of SProTrack. (a) Stage 1: a Spectral-Spatial AutoEncoder (SSAE) is trained self-supervisedly to encode hyperspectral frames into compact latent representations. (b) Stage 2: the frozen SSAE encoder generates latent spectral prompts, injected into a transformer-based RGB backbone for layer-wise modulation. (c) Spectral-Spatial block: a spectral branch models inter-band dependencies via Mamba and a spatial branch captures local patterns via convolution; their fused outputs, stacked across blocks, form the encoder, with a symmetric decoder for reconstruction. Detailed implementations are provided in the supplementary material.}
    \label{fig:overall}
\end{figure*}

\section{Method}
\label{sec:method}


\noindent\textbf{Problem setting.} 
Let $\{\boldsymbol{\mathcal{I}}_t\}_{t=1}^{T}$ denote a hyperspectral video of length $T$, where each frame $\boldsymbol{\mathcal{I}}_t \in \mathbb{R}^{B \times h \times w}$ contains $B$ spectral bands. Given the initial bounding box $\mathbf{b}_0 \in \mathbb{R}^4$, the objective is to estimate the target state $\mathbf{b}_t$ for each subsequent frame $t$.
Following the tracking-by-detection paradigm, each frame is represented by a \emph{search region} $\boldsymbol{\mathcal{X}}_t$ and a \emph{template set} $\boldsymbol{\mathcal{Z}}_t$ constructed from historical target appearances. The tracking model is formulated as:
$\mathcal{F}: \left(\boldsymbol{\mathcal{X}}_t, \boldsymbol{\mathcal{Z}}_t, \boldsymbol{\Theta}_{t-1}\right) \rightarrow \left(\mathbf{b}_t, \boldsymbol{\Theta}_t\right)$,
where $\boldsymbol{\Theta}_t$ denotes the temporal state encoding trajectory information across frames.
We adopt a transformer-based RGB tracker as the temporal backbone. The search and template images are divided into non-overlapping patches and embedded into token sequences $\boldsymbol{S}_x \in \mathbb{R}^{D \times N_x}$ and $\boldsymbol{S}_z \in \mathbb{R}^{D \times N_z}$. These tokens are concatenated with a temporal token $\boldsymbol{\Theta}_t$:
\begin{equation}
\boldsymbol{\Sigma}_0^t = \boldsymbol{S}_x \oplus \boldsymbol{S}_z \oplus \boldsymbol{\Theta}_t,
\end{equation}
and processed by a transformer encoder $\mathcal{T}$ with $L$ layers:
\begin{equation}
\boldsymbol{\Sigma}_l^t = \mathcal{T}_l(\boldsymbol{\Sigma}_{l-1}^t), \quad l = 1, \dots, L.
\end{equation}
The temporal token is updated and propagated across frames to maintain trajectory consistency.

\noindent\textbf{Overview.} As shown in \cref{fig:overall}, SProTrack decouples temporal modeling from spectral adaptation in two stages: a self-supervised SSAE learns compact spectral representations (\cref{sec:ssae}), which are then injected as layer-wise prompts into a temporal RGB tracking backbone (\cref{sec:prompting}). 

\subsection{Decoupled Multimodal Framework}

HOT involves modeling both temporal dynamics across frames and spectral characteristics across wavelength bands. Jointly learning these components from limited hyperspectral data often leads to overfitting due to data scarcity.
To address this limitation, we propose a \emph{decoupled multimodal framework} that separates temporal modeling from spectral representation learning, allowing each component to be optimized independently and more effectively.
The tracking process is decomposed into:
\begin{equation}
\begin{aligned}
\mathbf{p}_t = \mathcal{P}(\boldsymbol{\mathcal{L}}_t), \quad 
\mathbf{b}_t = \mathcal{T}_{\text{RGB}}\big(\boldsymbol{\Sigma}_t;\, \mathbf{p}_t\big),
\end{aligned}
\end{equation}
where $\mathcal{T}_{\text{RGB}}$ denotes a pretrained RGB tracking backbone and $\mathcal{P}$ denotes the prompt module that generates spectral prompts $\mathbf{p}_t$ from the latent representation $\boldsymbol{\mathcal{L}}_t$. This formulation explicitly decouples temporal modeling from spectral adaptation.
We use a pretrained transformer-based tracker to model temporal dynamics and keep its parameters frozen. This preserves strong trajectory modeling capability learned from large-scale RGB data and avoids overfitting under limited hyperspectral supervision.
Spectral information is captured independently through a compact latent representation learned from full hyperspectral frames. This representation encodes global inter-band correlations and serves as the basis for cross-modal adaptation.

Multimodal fusion is achieved via latent prompting, where the spectral representation is transformed into prompts that modulate intermediate features of the RGB backbone. This enables deep, layer-wise interaction between spectral and temporal information without altering the backbone architecture.
Decoupling representation learning from temporal modeling improves data efficiency and preserves pretrained temporal knowledge, treating spectral cues as complementary guidance rather than a replacement for RGB temporal modeling.

\subsection{Latent Spectral Representation Learning}
\label{sec:ssae}

\noindent\textbf{Spectral-spatial encoding.} To support the decoupled framework, we learn a compact and expressive latent representation from hyperspectral data in a self-supervised manner.
Given a hyperspectral frame $\boldsymbol{\mathcal{I}} \!\in\! \mathbb{R}^{B \times h \times w}$ with $B$ spectral bands, we aim to learn a low-dimensional latent representation $\boldsymbol{\mathcal{L}} \!\in\! \mathbb{R}^{C \times h \times w}$, where $C \!\ll\! B$. This representation preserves spatial structures and inter-band correlations while remaining efficient for downstream integration.
We adopt an \emph{SSAE} consisting of an encoder $\mathcal{E}$ and a decoder $\mathcal{D}$:
\begin{equation}
\boldsymbol{\mathcal{L}} = \mathcal{E}(\boldsymbol{\mathcal{I}}), \quad
\hat{\boldsymbol{\mathcal{I}}} = \mathcal{D}(\boldsymbol{\mathcal{L}}).
\end{equation}

The encoder jointly models spectral dependencies and spatial structures through a sequence of Spectral-Spatial Blocks (see \cref{fig:overall} (c)). Given feature maps $\boldsymbol{\mathcal{Z}} \in \mathbb{R}^{C \times H \times W}$, each block consists of two complementary branches.
The spectral branch captures long-range inter-band dependencies by partitioning features into spatial patches and reshaping them into spectral token sequences, which are processed by a sequence modeling module. The spatial branch applies convolutional operations to capture local spatial patterns independent of spectral ordering. The outputs of both branches are fused through a learnable projection, enabling joint spectral-spatial representation learning.
We instantiate the sequence modeling module using a Mamba block, which efficiently models long-range dependencies with linear complexity, making it suitable for high-dimensional hyperspectral data. The final latent representation is obtained via a $1\times1$ convolution followed by normalization.

\noindent\textbf{Learning objective.}
The SSAE is trained without annotations by reconstructing the input hyperspectral
image (see Fig.~2(a)). We combine a spatial-domain and a frequency-domain term ($\lambda_{f}=0.1$):
\begin{equation}
\mathcal{L}_{\text{SSAE}}
= \big\lVert \mathcal{I}-\hat{\mathcal{I}} \big\rVert_{1}
+ \lambda_{f}\,\Big\lVert\, \lvert\mathcal{F}_{b}(\mathcal{I})\rvert
- \lvert\mathcal{F}_{b}(\hat{\mathcal{I}})\rvert \,\Big\rVert_{1},
\label{eq:ssae}
\end{equation}
$\mathcal{F}_{b}(\cdot)$ is the orthonormal 2-D DFT taken over the two spatial
dimensions of each band independently; no transform is applied along the spectral
axis. Concretely, we compute the DFT of the input and of the reconstruction
separately, take the magnitude of each, and average the absolute difference over
all bands and frequency bins. Inter-band structure is modelled by the spectral
branch of the encoder rather than by this objective. The resulting latent representation captures global spectral structure from full,
unlabeled hyperspectral frames, and serves as a compact interface for the
subsequent multimodal adaptation via prompting.

\begin{table*}[tbp]
\caption{State-of-the-art comparison on HOT2020 and HOT2023. RGB trackers use RGB/FCI inputs, while hyperspectral methods use FCI + HSI. Best and second-best results are in \textbf{bold} and \underline{underline}. }
\vspace{-0.15cm}
\centering
\resizebox{0.85\textwidth}{!}{%
\begin{tabular}{l|l|cc|cc|cc|cc|cc}
\hline
& \multirow{3}{*}{Method} & \multicolumn{2}{c|}{HOT2020} & \multicolumn{8}{c}{HOT2023} \\
\cline{3-12}
& & \multicolumn{2}{c|}{RGB/HSI} & \multicolumn{2}{c|}{FCI/NIR} & \multicolumn{2}{c|}{FCI/RedNIR} & \multicolumn{2}{c|}{FCI/VIS} & \multicolumn{2}{c}{ALL} \\
\cline{3-12}
& & DP & AUC & DP & AUC & DP & AUC & DP & AUC & DP & AUC \\
\hline
\multirow{7}{*}{RGB trackers} & SeqTrack \cite{chen2023seqtrack} & 0.908 & 0.668 & 0.834 & 0.643 & 0.631 & 0.470 & 0.831 & 0.603 & 0.798 & 0.593 \\
& OSTrack \cite{ostrack} & 0.871 & 0.636 & 0.887 & 0.670 & 0.631 & 0.498 & 0.796 & 0.590 & 0.807 & 0.606 \\
& SMAT \cite{gopal2024separablesmat} & 0.831 & 0.600 & 0.866 & 0.669 & 0.448 & 0.335 & 0.744 & 0.531 & 0.737 & 0.550 \\
& AQATrack \cite{aqatrack} & 0.865 & 0.647 & 0.809 & 0.618 & 0.613 & 0.424 & 0.820 & 0.603 & 0.785 & 0.585 \\
& EVPTrack \cite{shi2024evptrack} & 0.870 & 0.649 & 0.843 & 0.644 & 0.574 & 0.424 & 0.801 & 0.590 & 0.785 & 0.585 \\
& ARTrackV2 \cite{bai2024artrackv2promptingautoregressivetracker} & 0.868 & 0.635 & 0.910 & 0.667 & 0.524 & 0.375 & 0.798 & 0.547 & 0.800 & 0.568 \\
& ODTrack \cite{odtrack} & 0.930 & \underline{0.693} & 0.924 & 0.694 & 0.606 & 0.475 & 0.859 & 0.653 & 0.839 & 0.641 \\
\hline
& MHT \cite{9950174mht} & 0.883 & 0.586 & 0.756 & 0.418 & 0.442 & 0.316 & 0.765 & 0.507 & 0.717 & 0.453 \\
& ViPT \cite{vipt} & 0.909 & 0.669 & 0.950 & 0.742 & 0.709 & 0.563 & 0.858 & 0.641 & 0.871 & 0.668 \\
& SEE-Net \cite{seenet} & \underline{0.933} & 0.666 & 0.785 & 0.465 & 0.491 & 0.359 & 0.813 & 0.593 & 0.763 & 0.519 \\
& SPIRIT \cite{spirit} & 0.924 & 0.678 & 0.838 & 0.623 & 0.504 & 0.381 & 0.822 & 0.608 & 0.786 & 0.585 \\
Hyperspectral & SiamCAT \cite{10474033siamcat} & 0.907 & 0.646 & 0.920 & 0.665 & 0.465 & 0.479 & 0.826 & 0.596 & 0.838 & 0.605 \\
trackers & MMF-Net \cite{li2024materialmmf} & 0.931 & 0.689 & 0.910 & 0.666 & 0.410 & 0.324 & 0.815 & 0.613 & 0.797 & 0.595 \\
& Trans-DAT \cite{10491347transdat} & 0.769 & 0.532 & 0.900 & 0.683 & 0.726 & 0.535 & 0.725 & 0.511 & 0.782 & 0.573 \\
& DaSSPNet \cite{dasspnet} & 0.916 & 0.680 & \underline{0.952} & \textbf{0.746} & \underline{0.731} & \textbf{0.587} & \underline{0.865} & \underline{0.654} & \underline{0.878} & \underline{0.678} \\
\hline
\multirow{1}{*}{Ours} 

& \textbf{SProTrack} & \textbf{0.952} & \textbf{0.704} & \textbf{0.970} & \underline{0.741} & \textbf{0.742} & \underline{0.583} & \textbf{0.886} & \textbf{0.665} & \textbf{0.906} & \textbf{0.697} \\

\hline
\end{tabular}
}
\vspace{-0.1cm}
\label{tab:comparison}
\end{table*}

\subsection{Latent Spectral Prompting}
\label{sec:prompting}
\noindent\textbf{Prompt generation.} We integrate spectral information into the tracking backbone through a latent prompting mechanism that enables efficient and deep multimodal fusion.
Given the latent representation $\boldsymbol{\mathcal{L}}_t \in \mathbb{R}^{C \times h \times w}$, we first crop the latent to search/template regions corresponding to their false color counterparts, then project it into the token space to obtain spectral prompts:
\begin{equation}
\mathbf{p}_0^t = \phi(\text{cropped}(\boldsymbol{\mathcal{L}}_t)),
\end{equation}
where $\phi(\cdot)$ denotes a learnable projection function. 

\noindent\textbf{Prompt injection.} The resulting tokens $\mathbf{p}_0^t \in \mathbb{R}^{D \times N}$ encode compact spectral cues aligned with the RGB backbone. To enable deep fusion, the prompts are progressively computed and injected into each transformer layer. Since the SSAE operates on the full hyperspectral frame and preserves spatial correspondence with its input, cropping the latent by the same box coordinates used for the RGB search and template regions yields features that align with the corresponding RGB token grids. At layer $l$, the prompts are updated as:
\begin{equation}
\mathbf{p}_l^t = \mathcal{P}_l\big(\boldsymbol{\Sigma}_{l-1}^t, \mathbf{p}_{l-1}^t\big),
\end{equation}
where $\mathcal{P}_l$ denotes a lightweight learnable prompt module that
maps the previous backbone features and prompts to the layer-$l$ prompt
of dimension $D\times N$. The prompt is then used to modulate the
corresponding backbone features:
\begin{equation}
\boldsymbol{\Sigma}_l^t = \mathcal{T}_l\big(\boldsymbol{\Sigma}_{l-1}^t + \mathbf{p}_l^t\big).
\end{equation}
Here, the addition is element-wise over corresponding spatial tokens, while the temporal token is kept unchanged. This layer-wise interaction allows spectral information to influence representations at multiple semantic levels.

\noindent\textbf{Temporal consistency.}
We preserve temporal modeling by maintaining a temporal token $\boldsymbol{\Theta}_t \in \mathbb{R}^{D}$, which encodes the target trajectory across frames. The input sequence is constructed as:
\begin{equation}
\boldsymbol{\Sigma}_0^t = \boldsymbol{S}_x^t \oplus \boldsymbol{S}_z^t \oplus \boldsymbol{\Theta}_t,
\end{equation}
and the temporal state is propagated as:
\begin{equation}
\boldsymbol{\Theta}_{t+1} = \boldsymbol{\Theta}_L^t + \boldsymbol{\Theta}_{\text{init}},
\end{equation}
where $\boldsymbol{\Theta}_{\text{init}} \in \mathbb{R}^{D}$ is a learnable initialization token. By keeping the temporal update unchanged, the model retains robust trajectory modeling while incorporating spectral guidance.
Latent spectral prompting enables implicit multimodal fusion through layer-wise feature modulation, allowing efficient adaptation to hyperspectral inputs while preserving pretrained temporal knowledge.

\subsection{Training and Adaptation}

We adopt a two-stage training strategy to optimize the proposed framework in a data-efficient manner. 

\noindent\textbf{Stage 1: spectral representation learning.}
The SSAE is trained in a fully self-supervised manner using hyperspectral data without annotations. We use the loss defined in \cref{eq:ssae}. The SSAE is trained only on the training split of each benchmark to avoid any test-time information leakage.

\noindent\textbf{Stage 2: multimodal tracking.}
The tracking model is initialized with the pretrained SSAE encoder and a pretrained RGB backbone (see \cref{fig:overall} (b)). The backbone parameters are frozen to preserve temporal modeling capability, 
while only the prompt generation modules, hyperspectral projection
layers, and LoRA adapters are optimized.
To improve parameter efficiency, we use LoRA to selected layers (e.g., projection-in and projection-out layers implemented as fully connected layers, as shown in \cref{fig:overall} (c)) in the spectral encoder, enabling effective task adaptation with minimal additional parameters. 
The overall tracking objective is defined as:
\begin{equation}
\mathcal{L} = L_{\text{cls}} + \lambda_1 L_1 + \lambda_2 L_{\text{GIoU}},
\end{equation}
where $L_{\text{cls}}$ is the classification loss, $L_{1}$ and $ L_{\text{GIoU}}$, with corresponding scales $\lambda_1=5$ and $\lambda_2=2$, denote bounding box regression losses following ODTrack\cite{odtrack} settings.
This training strategy uses self-supervised spectral learning and lightweight multimodal adaptation to achieve strong generalization under limited hyperspectral supervision.

\section{Experiment}
\label{sec:experiment}

\subsection{Datasets and Setups}

\noindent\textbf{Datasets.}
We evaluate on: 
\textit{(i) HOT2020} \cite{xiong2020material} contains 40 training and 35 testing sequences, all captured in visible (VIS) spectral domain. \textit{(ii) HOT2023} \cite{xiong2023hyperspectral} extends this setting to multiple spectral domains, including VIS, near-infrared (NIR), and red-to-near-infrared (RedNIR), providing a more challenging and diverse evaluation scenario.

\noindent\textbf{Setups.}
SProTrack is trained on a single NVIDIA RTX 5090 GPU, using ODTrack Base-Fulldata-300ep~\cite{got,muller2018trackingnet,lasot} as the frozen RGB foundation. For each spectral domain, we pretrain a separate SSAE for 150 epochs (batch size 16, patch size $p\!=\!8$) using AdamW (initial lr $8\!\times\!10^{-5}$, weight decay $1\!\times\!10^{-4}$, cosine annealing), on full hyperspectral images without bounding-box annotations, with random cropping, rotation, and horizontal flipping as augmentation. For tracking, we optimize only the prompt modules, hyperspectral projection layers, and LoRA on SSAE, while keeping the backbone frozen; each epoch uses 7{,}200 sampled pairs (batch size 16) with three $384\!\times\!384$ search and three $192\!\times\!192$ template images sampled within max gap 400 frames. Here the three search frames form a temporally adjacent clip sharing one template set, processed autoregressively so the temporal token propagates across them, following ODTrack\cite{odtrack}. On VIS, we train 60 epochs with a step schedule (lr decayed $10\times$ after 50 epochs). On NIR and RedNIR, where data is scarcer, we train 40 epochs with a constant learning rate to prevent overfitting.

\noindent\textbf{Metrics.} We follow the standard one-pass evaluation protocol, where the tracker processes each sequence sequentially without re-initialization. Performance is evaluated using: (i) the area under the success curve (AUC), computed from the intersection-over-union between predicted and ground-truth bounding boxes, and (ii) distance precision (DP), defined as the percentage of frames where the center location error is within a 20-pixel threshold.

\begin{figure}[tbp]
    \centering
    \includegraphics[width=0.95\linewidth]{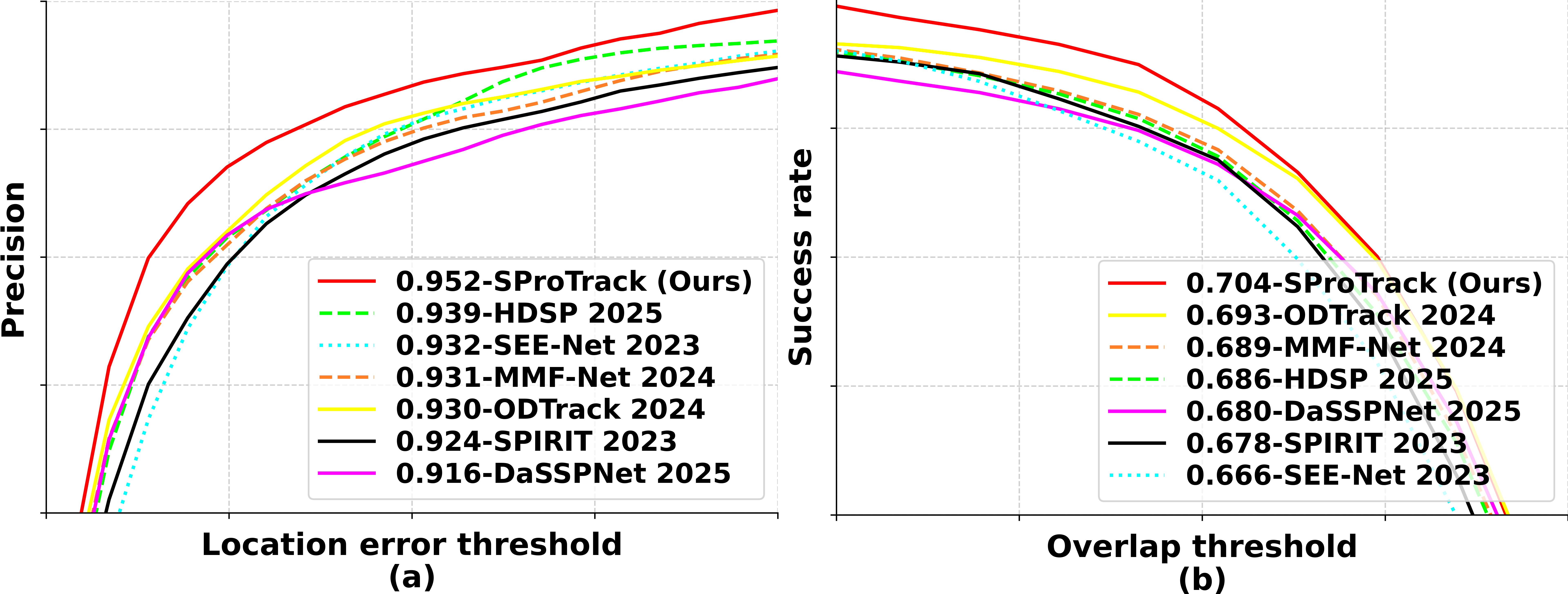}
    \vspace{-0.2cm}
    \caption{Precision and success curves on HOT2020, where (a) shows the precision plot and (b) shows the success plot.}
    \label{fig:success_precision_plot}
\end{figure}

\subsection{Comparison with State-of-the-Art}

\cref{tab:comparison} presents a comparison with state-of-the-art RGB and hyperspectral trackers on HOT2020 and HOT2023. SProTrack achieves the best DP across all domains and the best overall (ALL) AUC. On HOT2020, our method attains 0.952 DP and 0.704 AUC, surpassing the strongest RGB baseline ODTrack\cite{odtrack} and recent hyperspectral methods such as DaSSPNet\cite{dasspnet} (see \cref{fig:success_precision_plot}). We trained with 5 random seeds and observed a standard deviation of 0.0018. On HOT2023, we achieve 0.906 DP and 0.697 AUC on the aggregated setting, improving over the second-best by a clear margin. Notably, SProTrack stays strong even on the challenging NIR and RedNIR domains where many methods degrade, owing to globally learned spectral representations that capture material properties beyond individual bands and are integrated via layer-wise prompting. 

\noindent\textbf{RGB trackers.} When applied to hyperspectral data via false-color images, RGB trackers exhibit strong performance in the VIS domain but degrade in NIR and RedNIR settings due to the domain gap. For example, ODTrack\cite{odtrack} achieves competitive results on VIS but drops noticeably on RedNIR. In contrast, SProTrack consistently improves performance across all domains, indicating that latent spectral prompting effectively bridges the spectral gap without sacrificing temporal modeling capability.

\noindent\textbf{Hyperspectral trackers.} Compared to dedicated hyperspectral trackers, our method achieves the best overall performance, with superior or competitive results across individual domains. Notably, while DaSSPNet\cite{dasspnet} achieves strong results in certain domains, our method provides more consistent gains across VIS, NIR, and RedNIR, leading to the best overall performance. This highlights the advantage of decoupling spectral representation learning from temporal modeling and integrating them through latent prompting.



\begin{table}[tbp]
\caption{Comparison of efficiency and tracking performance (MACs, FPS, parameters, AUC).}
\vspace{-0.15cm}
\centering
\resizebox{0.95\linewidth}{!}{
\begin{tabular}{l|cccc}
\hline
Method & MACs (G) & FPS & Params (M) & AUC \\
\hline
ODTrack \cite{odtrack} & 103.3 & 74.8 & 92.83 & 0.693 \\
DaSSPNet \cite{dasspnet}& 25.9 & 43.8 & 95.56 & 0.680 \\
SProTrack (ours) & 145.14 & 43.1 & 111.5 & 0.704 \\
\hline
\end{tabular}}
\label{tab:efficiency}
\vspace{-0.2cm}
\end{table}

\subsection{Analysis and Discussion}


\noindent\textbf{Computational efficiency.} We evaluate efficiency in terms of MACs, inference speed (FPS), and parameter count, as summarized in Tab.~\ref{tab:efficiency}. SProTrack runs at 43.1 FPS on a single RTX 5090 GPU, satisfying real-time requirements and on par with DaSSPNet\cite{dasspnet} (43.8 FPS) despite a substantially higher MACs count. This apparent gap arises because MACs measure theoretical operations, whereas wall-clock speed also reflects hardware efficiency: our Mamba-based spectral branch attains higher throughput than the Conv3D operations common in prior hyperspectral trackers, and the SSAE latent for the template is computed once and cached across frames rather than recomputed. As a result, the spectral modeling adds little to the per-frame critical path. While the total parameter count (111.5M) is higher than prior methods, the backbone is frozen and only 0.895M parameters (0.80\%) are actually trainable, so cross-modal adaptation is achieved by tuning under 1\% of the network. The higher MACs stem mainly from the transformer-based ODTrack\cite{odtrack} backbone, which operates on larger input resolutions and multiple reference frames. Beyond that the additional cost yields consistent accuracy gains: +1.1\% AUC over ODTrack\cite{odtrack} and +2.4\% over DaSSPNet\cite{dasspnet}, demonstrating a favorable accuracy-efficiency trade-off.

\begin{table*}[tbp]
\caption{Attribute-based AUC comparison on the HOT2020 benchmark over 11 challenging attributes: background clutter (BC), deformation (DEF), fast motion (FM), illumination variation (IV), in-plane rotation (IPR), low resolution (LR), motion blur (MB), occlusion (OCC), out-of-plane rotation (OPR), out-of-view (OV), and scale variation (SV). The best and second results are in \textbf{bold} and \underline{underline}.}
\centering
\resizebox{0.8\textwidth}{!}{\begin{tabular}{l|ccccccccccc}
\hline
Method      & BC    & DEF   & FM    & IV    & IPR   & LR    & MB    & OCC   & OPR   & OV    & SV \\
\hline
BAE-Net~\cite{baenet}     & 0.651         & 0.680         & 0.608         & 0.525         & 0.699         & 0.489         & 0.593 & 0.555 & 0.701 & 0.517 & 0.604\\
SEE-Net~\cite{seenet}     & 0.705         & 0.710         & 0.711         & 0.566         & 0.742         & 0.626         & 0.702 & 0.622 & 0.729 & 0.633 & 0.649\\
SPIRIT~\cite{spirit}      & \textbf{0.715} & 0.723        & 0.715         & 0.567         & 0.760         & 0.629         & 0.723 & 0.643 & 0.751 & 0.689 & 0.665\\
ViPT~\cite{vipt}        & 0.633         & 0.764         & 0.685         & 0.642         & 0.743         & 0.637         & \textbf{0.733} & 0.667 & 0.749 & \textbf{0.822} & 0.673\\
DaSSPNet~\cite{dasspnet}    & 0.655         & 0.765        & 0.708         & \underline{0.659}         & \underline{0.762}         & 0.655         & 0.724 & 0.683 & 0.763 & \underline{0.815} & 0.689\\
HDSP~\cite{hdsp}        & 0.660         & \underline{0.769}  & \underline{0.721}         & 0.655         & \textbf{0.767}         & \textbf{0.672}         & 0.718 & \underline{0.684} & \underline{0.772} & 0.767 & \underline{0.695}\\
\hline
\textbf{SProTrack (ours)}        & \underline{0.714} & \textbf{0.782} & \textbf{0.727} & \textbf{0.665} & 0.740 & \underline{0.670} & \underline{0.730} & \textbf{0.694} & \textbf{0.773} & 0.724 & \textbf{0.704}\\
\hline
\end{tabular}}
\label{tab:attribute}
\vspace{-0.1cm}
\end{table*}

\noindent\textbf{Attribute-based analysis.} We report attribute-based AUC on HOT2020 across 11 attributes in \cref{tab:attribute}. SProTrack ranks first on 6 attributes (DEF, FM, IV, OCC, OPR, SV) and second on 2 (BC, MB). Gains concentrate on two groups: appearance-variation attributes (DEF, OCC, IV), where latent spectral representations capture material properties that stay stable when RGB cues become unreliable; and motion-dynamics attributes (FM, OPR, SV), where the inherited temporal priors from the frozen RGB backbone maintain trajectory continuity under rapid change. Our method slightly underperforms on OV and IPR by DaSSPNet~\cite{dasspnet} and SPIRIT~\cite{spirit}, whose explicit domain-adaptation modules align spectral distributions across conditions, an advantage under extreme viewpoint or visibility changes.

\begin{figure}[tbp]
    \centering
    \includegraphics[width=0.9\columnwidth]{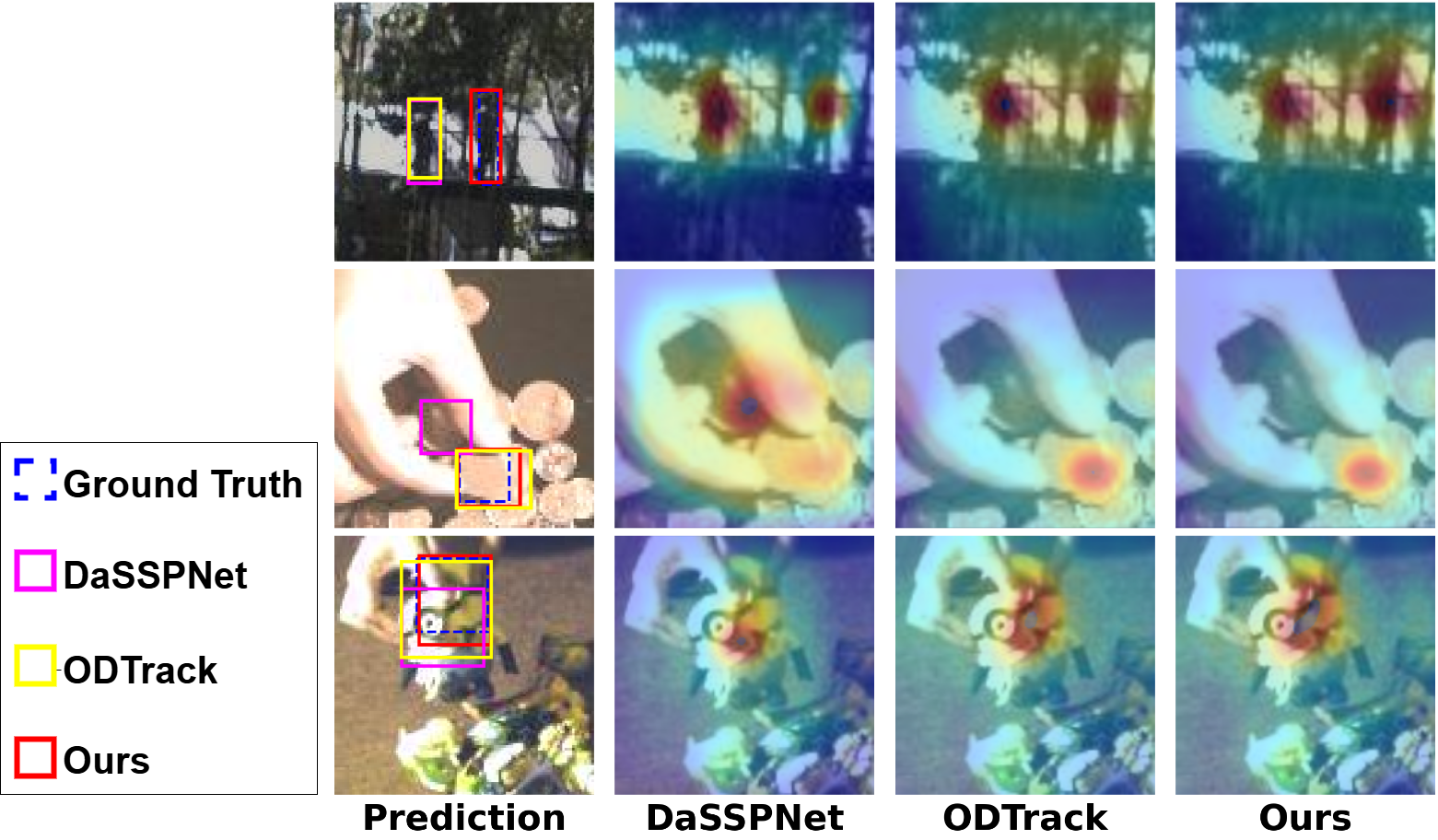}
    \vspace{-0.2cm}
    \caption{Qualitative comparison on three challenging HOT2020 sequences. Our method yields more concentrated responses under occlusion, distractors, and appearance ambiguity.}
    \label{fig:qualitative}
\end{figure}

\noindent\textbf{Qualitative results.} \cref{fig:qualitative} compares three challenging HOT2020 sequences. 
Across all cases, our method produces more concentrated, well-centered responses, enabling accurate localization where competing methods drift. Under partial occlusion with complex illumination (row 1), baselines yield diffuse responses with background distractions, while ours stays compact. With a nearby similar distractor (row 2), competing methods activate both objects, whereas ours gives a sharp unimodal response. Under occlusion (row 3), most methods shift toward the occluder, while ours preserves activation on the true target. Further qualitative results, failure cases, statistical analysis, and limitations are provided in the supplementary material.


\begin{figure}[tbp]
    \centering
    \includegraphics[width=0.85\columnwidth]{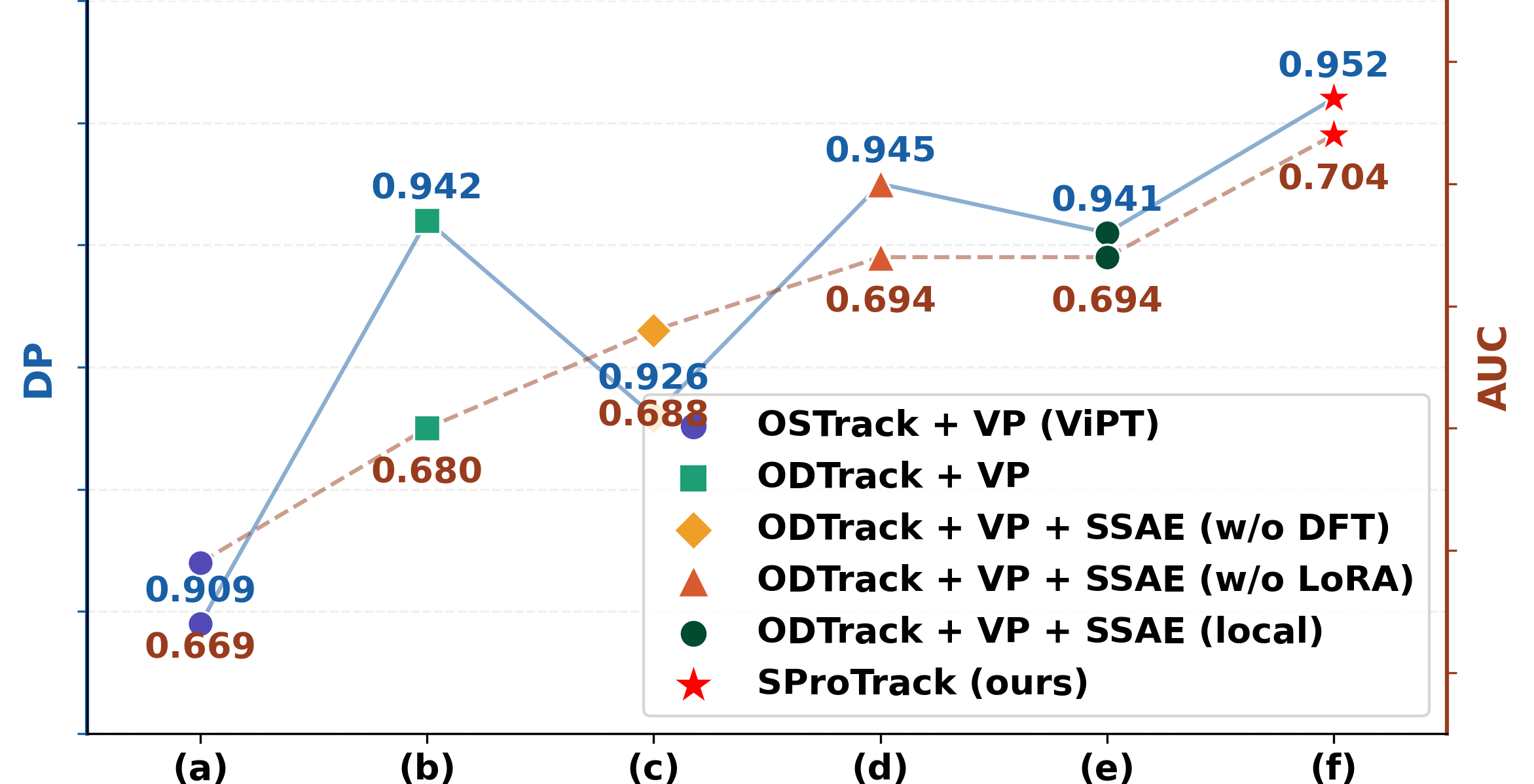}
    \vspace{-0.2cm}
    \caption{Ablation study on HOT2020 illustrating the impact of progressively adding key components. VP denotes visual prompting. Each point corresponds to a model variant, evaluated by distance precision (DP) and success (AUC). }
    \label{fig:ablation_components}
\end{figure}

\subsection{Ablation Study}
We ablate each component on HOT2020 (\cref{fig:ablation_components}, \cref{tab:ssae}).

\noindent\textbf{Temporal and spectral modeling are complementary. 
} 
The 0.693 AUC reported for ODTrack in \cref{tab:comparison} is the plain ODTrack baseline, whereas the 0.680 AUC in \cref{fig:ablation_components}(b) is ODTrack with visual prompting (VP); the latter is therefore the matched baseline for evaluating the additional effect of SSAE prompting.
Moving from OSTrack\cite{ostrack}+VP (ViPT) to ODTrack\cite{odtrack}+VP raises AUC by 1.1\% (\cref{fig:ablation_components}(a) vs. (b)). 
Although this configuration performs worse than the plain version in \cref{tab:comparison}, likely due to overfitting, 
the use of pretrained temporal priors is deliberate.
Under severe data scarcity, training a strong temporal tracker from scratch is challenging, so our model reuses pretrained RGB temporal knowledge. 
On top of this backbone, SSAE prompting adds a further 2.4\% AUC (\cref{fig:ablation_components}(b) vs. (f)), exceeding the gain from the backbone change itself. The two components are thus complementary: reused temporal priors handle trajectory modeling, while spectral prompting supplies discriminative cues unavailable from RGB pretraining alone.

\noindent\textbf{Global vs. local spectral context.} To isolate the role of full-frame modeling, we train an SSAE variant (\cref{fig:ablation_components}(e)) using only the hyperspectral content within the (jittered) search-crop region, matching the local-region inputs of prior prompt-based methods. Compared to no SSAE (\cref{fig:ablation_components}(b)), this local variant yields almost no improvement, indicating that injecting spectral features from local crops alone adds little. In contrast, our full-frame SSAE (\cref{fig:ablation_components}(f)) clearly surpasses both (+1.1\% DP, +1.0\% AUC). This shows that the benefit of SSAE comes not from adding spectral features per se, but from encoding \emph{global}, scene-level context that local crops cannot capture, directly validating our full-frame design.

\begin{figure}[tbp]
    \centering
    \includegraphics[width=0.95\columnwidth]{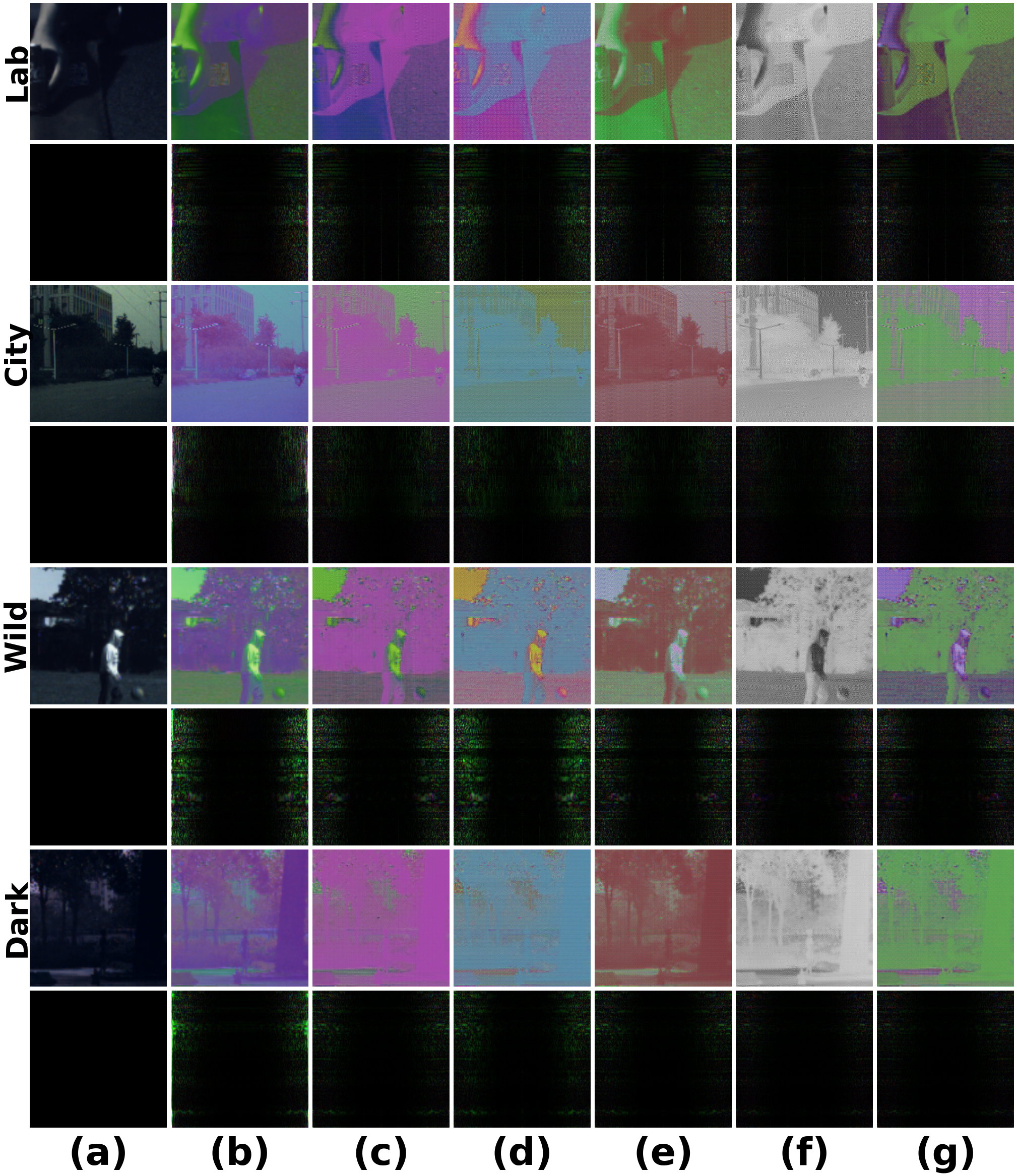}
    \vspace{-0.2cm}
    \caption{SSAE ablation on HOT2020. Each column shows the reconstructed latent and its frequency-domain error map (darker means lower error). (a) input false-color image and reference (zero-error) map. (b) -- (g) SSAE configurations: (b) CNN-only; (c) dual-branch, 2 blocks; (d) 1 latent channel; (e) without frequency loss; (f) standard SSAE (dual-branch, 3 blocks, 3 channels); (g) 5 channels. Settings match \cref{tab:ssae}.}
    \label{fig:ablation_ssae}
\end{figure}

\noindent\textbf{Frequency-domain supervision.} Removing
DFT degrades
precision from $0.952$ to $0.926$~DP (\cref{fig:ablation_components}(c)). Because DFT
operates on band-wise magnitude spectra under an $\ell_{1}$ norm, it emphasizes
the low-energy high-frequency components that a spatial reconstruction loss
under-weights. The resulting sharper reconstruction of edges and texture
(\cref{fig:ablation_ssae}(e) vs. (f)) benefits center-accurate
localization, which DP measures directly, more than it benefits overlap-based
AUC.

\noindent\textbf{Parameter-efficient adaptation.} Removing LoRA (\cref{fig:ablation_components}(d)) lowers both metrics (0.945 DP, 0.694 AUC); the lightweight adaptation recovers the gap with negligible overhead, indicating it suffices to align spectral representations with the tracking objective without overfitting.

\noindent\textbf{Scalability.} The decoupled design lets the SSAE leverage extra unlabeled data: pretraining on the larger HOT2023 set and transferring to HOT2020 yields further gains (+0.4\% DP, +0.2\% AUC) without any additional annotations.

\begin{table}[tbp]
\caption{Comparison of configurations of SSAE.  \label{tab:ssae}}
\centering
\resizebox{0.9\linewidth}{!}{
\begin{tabular}{ccccc}
\hline
Block & \#Block & \#Channel & Validation Loss & AUC\\
\hline
CNN & 3 & 3 & 0.0613 & 0.694 \\
Dual & 2 & 3 & 0.0289 & 0.691 \\
Dual & 3 & 1 & 0.0397 & 0.693\\
Dual & 3 & 3 & \underline{0.0277} & \underline{0.704}\\
Dual & 3 & 5 & \textbf{0.0255} & \textbf{0.707}\\
\hline
\end{tabular} }
\end{table}
\noindent\textbf{SSAE design.} We analyze the SSAE design in \cref{tab:ssae} and \cref{fig:ablation_ssae}. 
The CNN-only encoder
produces structured artifacts near image boundaries, while the dual-branch design better captures global spectral-spatial structure. Increasing the latent dimensionality from 1 to 3 substantially improves reconstruction and tracking, while the gain from 3 to 5 channels is marginal (+0.003 AUC) despite the increased latent dimensionality. We therefore adopt the 3-channel configuration as a more compact accuracy-efficiency trade-off. Removing frequency supervision increases residual high-frequency error, further supporting the use of the DFT loss.


\noindent\textbf{Generalization across backbones.} To verify that SProTrack is not tied to a particular architecture, we also evaluate it on OSTrack~\cite{ostrack} and SeqTrack~\cite{chen2023seqtrack}. For each backbone, we compare the original tracker with its SProTrack-enhanced counterpart under the same protocol. As shown in \cref{tab:cross_backbone}, SProTrack's mechanism consistently improves AUC and DP on both HOT2020 and HOT2023 across all backbones, demonstrating that the performance gains are not specific to ODTrack~\cite{odtrack}. 
SProTrack is not restricted to conventional RGB trackers and could, in principle, be adapted to zero-shot trackers such as CoTracker\cite{karaev2024cotracker} and CoTracker 3\cite{karaev2025cotracker3} by injecting spectral prompts into their intermediate visual representations. However, adapting its point-tracking formulation to the bounding-box tracking setting of HOT would require additional design.

\begin{table}[t]
\centering\small\setlength{\tabcolsep}{4pt}
\caption{Cross-backbone generalization. Each backbone uses SProTrack under the same protocol. 
}
\vspace{-0.1cm}
\label{tab:cross_backbone}
\resizebox{0.9\linewidth}{!}{\begin{tabular}{ll cc cc}
\toprule
\multirow{2}{*}{Backbone} & \multirow{2}{*}{Paradigm} & \multicolumn{2}{c}{HOT2020} & \multicolumn{2}{c}{HOT2023 ALL} \\
\cmidrule(lr){3-4} \cmidrule(lr){5-6}
 & & DP & AUC & DP & AUC \\
\midrule
OSTrack~\cite{ostrack}          & \multirow{2}{*}{one-stream}     & 0.871 & 0.636 & 0.807 & 0.606 \\
\quad + SProTrack               &                                 & \textbf{0.922} & \textbf{0.665} & \textbf{0.875} & \textbf{0.669} \\
\midrule
SeqTrack~\cite{chen2023seqtrack} & \multirow{2}{*}{autoregressive} & 0.908 & 0.668 & 0.798 & 0.593 \\
\quad + SProTrack               &                                 & \textbf{0.923} & \textbf{0.677} & \textbf{0.878} & \textbf{0.669} \\
\midrule
ODTrack~\cite{odtrack}          & \multirow{2}{*}{temporal}       & 0.930 & 0.693 & 0.839 & 0.641 \\
\quad + SProTrack               &                                 & \textbf{0.952} & \textbf{0.704} & \textbf{0.906} & \textbf{0.697} \\

\bottomrule
\end{tabular}}
\vspace{-0.3cm}
\end{table}
\section{Conclusion}
\label{sec:conclusion}

We present SProTrack, a decoupled framework that learns spectral representations independently of temporal modeling. It achieves state-of-the-art results across HOT2020 and HOT2023, with gains driven primarily by latent spectral prompting and effective spectral fusion rather than model capacity. 
Future work will explore band-agnostic spectral encoders, applying the learned spectral prompts to zero-shot trackers such as CoTracker\cite{karaev2024cotracker}, CoTracker 3\cite{karaev2025cotracker3}, and SAM2\cite{ravi2024sam2segmentimages}, and explicit re-detection to address prolonged-occlusion and out-of-view failures.

\section*{Acknowledgments}

Wangshu Cai is supported by the Griffith University International Postgraduate Research Scholarship and the Griffith University Postgraduate Research Scholarship.
This work was supported in part by the ARC Grant
DP250102625 `Spectral-spatial-temporal object tracking in hyperspectral videos'.

{
    \small
    \bibliographystyle{ieeenat_fullname}
    \bibliography{main}
}

\clearpage

\appendix

\newcommand{\lei}[1]{{\color{red} #1}}




\maketitlesupplementary

\section{Dataset Details}
\begin{table}[h]
\centering
\caption{Dataset statistics. \label{tab:dataset_statistics}}
\begin{tabular}{l|cccccc}
\hline
Dataset & Domain & Split & Band & \#Seqs & \#Images \\
\hline
HOT2020 & VIS & Train & 16 & 40 & 12,887 \\
HOT2020 & VIS & Test & 16 & 35 & 16,461 \\
\hline
HOT2023 & VIS & Train & 16 & 55 & 19,251 \\
HOT2023 & VIS & Test & 16 & 46 & 20,074 \\
HOT2023 & NIR & Train & 25 & 40 & 8,239\\
HOT2023 & NIR & Test & 25 & 30 & 4,872 \\
HOT2023 & RedNIR & Train & 15 & 15 & 6,412 \\
HOT2023 & RedNIR & Test & 15 & 11 & 3,500  \\
\hline
\end{tabular}
\end{table}
\begin{figure*}[tbp]
    \centering
    \includegraphics[width=\textwidth]{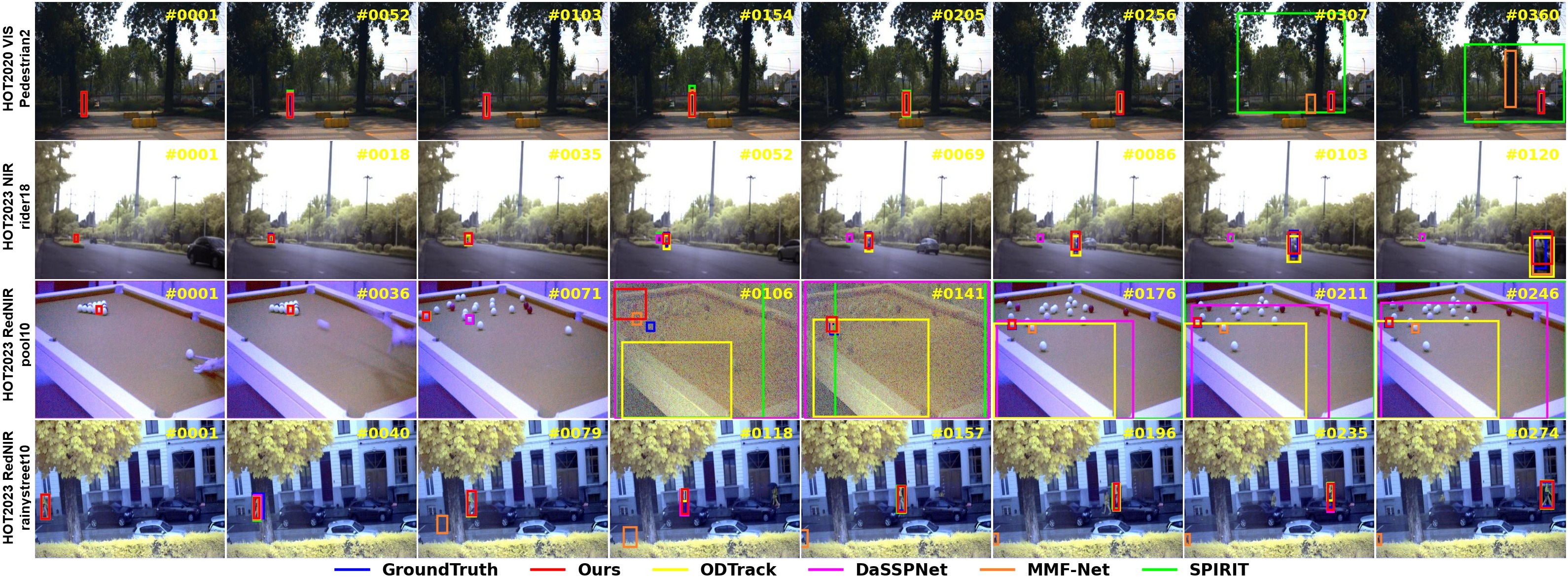}
    \caption{Additional qualitative comparison on four
        challenging sequences. The dataset name and sequence names are labeled on the left. Our method produces more concentrated responses and more accurate localization under occlusion, distractors, and bad lighting.}
    \label{fig:supplementary_qualitative}
\end{figure*}
We summarize the statistics of the hyperspectral object tracking datasets used in this paper in \cref{tab:dataset_statistics}. Both HOT2020 and HOT2023 contain diverse scenes including laboratories, urban streets, and outdoor environments, with image resolutions varying up to 1080p.

A notable imbalance exists across spectral domains. The VIS domain provides 55 training sequences with 19,251 frames in HOT2023, whereas NIR offers 40 sequences (8,239 frames) and RedNIR contains only 15 sequences (6,412 frames). This data scarcity in NIR and RedNIR domains directly impacts the SSAE's ability to learn robust spectral representations, as the self-supervised reconstruction objective benefits from diverse training samples. Furthermore, the NIR (25 bands) and RedNIR (15 bands) domains capture wavelength ranges that differ substantially from the visible spectrum on which the RGB foundation model was pretrained. This introduces a mismatch to VIS, making cross-modal adaptation inherently more challenging. These factors jointly explain the performance variation across domains observed in the main paper (\cref{tab:comparison}): our method achieves the largest improvements in VIS where sufficient training data is available, while gains in RedNIR and NIR are more modest due to data scarcity, different false-color quality, and the greater spectral distance from the RGB pretraining domain.


\section{Additional Qualitative Results}
 
We provide additional qualitative comparisons on difficult sequences across different spectral domains in \cref{fig:supplementary_qualitative}. 

Results further confirm that SProTrack maintains robust tracking under diverse spectral conditions. Comparing with DaSSPNet\cite{dasspnet}, our method provides more robustness to small objects (row~1 and row~2), distractors (row~3), and occlusion (row~1 and row~4). Comparing with our backbone ODTrack\cite{odtrack}, our method is more accurate, showing more concentration of scores to the objects of interest.

\section{Implementation Details}
\label{sec:supp_efficiency}
 
This section provides our implementation details.

\noindent\textbf{SSAE.}
The encoder maps the $B$ raw bands to $64$ channels with a $1{\times}1$
convolution, applies three dual-branch blocks, and projects to the 3-channel
latent with a second $1{\times}1$ convolution; the decoder mirrors this
structure. Each dual-branch block contains (i) a spatial branch, a depthwise
$3{\times}3$ convolution with $\text{groups}\!=\!64$, and (ii) a spectral-spatial
branch that tokenises the feature map into $8{\times}8$ patches, projects each
flattened patch ($8\times8\times64 \!=\! 4096$) to $512$ with a linear layer,
applies Mamba ($d_{\text{model}}{=}512$, $d_{\text{state}}{=}16$,
$d_{\text{conv}}{=}4$) along the resulting spatial token sequence, and projects
back to $4096$. The two branches are concatenated and fused by a $1{\times}1$
convolution ($128\rightarrow64$). $B$ is $16$ for HOT2020 and HOT2023-VIS,
$25$ for HOT2023-NIR and $15$ for HOT2023-RedNIR, corresponding to their band numbers.

\noindent\textbf{Prompt branch.}
We follow the ViPT-deep design with one prompt block per transformer layer ($12$ in total). Each block reduces the RGB and latent token maps from $768$ to $8$ channels with two $1{\times}1$ convolutions and projects the fused result back to $768$ with a third $1{\times}1$ convolution; a learnable scalar smooths the fovea weighting. The latent is embedded by a separate patch embedding of shape $768\times3\times16\times16$, and each layer carries its own LayerNorm ($12\times768$).


\noindent\textbf{LoRA.} LoRA is applied only to the two linear layers of each dual-branch block in the autoencoder encoder ($6$ layers, $12$ tensors), with rank $r=2$, $\alpha=4$, and dropout $0.1$. The tracker backbone and head contain no LoRA and remain frozen. With a $3$-channel latent, the model has 894,924 trainable parameters in total, comprising 55,296 LoRA parameters, 249,036 prompt-module parameters, and 590,592 projection parameters.

\noindent\textbf{Tracker and optimisation.}
The base tracker is ODTrack with a ViT-B/16 pre-trained backbone, candidate elimination at layers $\{3,6,9\}$ with keep ratio $0.7$, search region $384{\times}384$ and three $192{\times}192$ template frames. We use AdamW ($\text{wd}=10^{-4}$), a step schedule with decay $0.1$, a backbone learning-rate multiplier of $0.1$, drop-path $0.1$, and loss weights $\lambda_1 =5$, $\lambda_2 =2$. All runs use $7{,}200$ sampled pairs per epoch.

\begin{table}[t]\centering
\caption{Per-dataset training settings.}
\label{tab:hparams}
\setlength{\tabcolsep}{4pt}\small
\begin{tabular}{lccccccc}
\toprule
Dataset & Batch & Epochs & LR drop & LR & \\
\midrule
HOT2020-VIS          & 16 & 60 & 50 & $8\times10^{-5}$  \\
HOT2023-VIS     & 16 & 60 & 50 & $8\times10^{-5}$   \\
HOT2023-NIR     & 16 & 40 & -- & $8\times10^{-5}$  \\
HOT2023-RedNIR  & 16 & 40 & -- & $8\times10^{-5}$ \\
\bottomrule
\end{tabular}
\end{table}

\noindent\textbf{Frequency loss implementation.} DFT
is evaluated in \texttt{fp32}: inputs are cast from \texttt{fp16} before the
transform to avoid overflow in the accumulation. We use
\texttt{torch.fft.fft2} with \texttt{norm="ortho"}, which applies the transform
over the two trailing (spatial) axes of each band and yields the orthonormal
scaling under which Parseval's identity would hold for the residual spectrum.
Both loss terms use mean reduction. 

\section{Efficiency Details}

This section provides memory and training-cost measurements complementing the efficiency analysis in the main paper.

\noindent\textbf{Parameter efficiency.}
Although the full model contains 111.46M parameters, only 0.895M (0.80\%) are trainable, while the remaining 99.2\% (110.56M) are frozen. The trainable parameters consist solely of (i) LoRA modules inserted into the projection-in/projection-out layers of the SSAE, and (ii) the prompt modules 
and projection layers. The backbone, SSAE encoder, and prediction head are entirely frozen. 
We clarify the role and scope of LoRA, as it bears on our decoupling claim. The temporal axis is fully decoupled: the entire temporal backbone and prediction head are frozen, and no temporal parameter is updated by hyperspectral supervision. On the spectral side, the SSAE is pretrained self-supervised and its globally learned spectral representation is not re-trained for tracking; LoRA is applied only as a lightweight, task-conditioned readout on the SSAE projection layers, whose purpose is to extract subject-relevant spectral cues from the otherwise scene-level representation and route them into the prompt. Decoupling therefore holds in the sense that matters: temporal modeling and spectral representation learning are optimized in separate stages and never jointly, with only a sub-0.5M readout bridging them.

\noindent\textbf{Memory usage.}
At inference (batch size 1, no gradients, fp32), the model requires only 0.76~GiB of peak allocated memory and 0.94~GiB reserved; including the CUDA context ($\sim$0.5~GiB), the process occupies roughly 1.3~GiB. The model is therefore extremely lightweight to deploy, fitting comfortably on any modern GPU. During training (batch size 16, forward + backward + optimizer step), peak allocated memory is 15.24~GiB and peak reserved is 23.25~GiB. The footprint is dominated by activations rather than parameters or optimizer state: since only 0.80\% of parameters are trainable, the AdamW optimizer states and gradients are negligible, and the 15.24~GiB of allocated memory is almost entirely forward activations at batch size 16. We adopt a per-frame autoregressive backward pass that releases activations frame by frame, following ODTrack\cite{odtrack}, avoiding the roughly 3× activation memory that retaining all three search frames' computation graphs simultaneously would incur.
 
\noindent\textbf{Training cost.}
SProTrack is trained in two stages, both lightweight and run on a single GPU (\cref{tab:supp_training}). Stage~1 pretrains the SSAE in a fully self-supervised manner without annotations, at 2.2~minutes per epoch ($\sim$2.88 batches/s) for 150 epochs. Stage~2 trains only the prompt and projection modules for tracking, at 13.9~minutes per epoch (batch size 16, 7{,}200 sampled pairs per epoch, using three $384\times384$ search frames and three $192\times192$ template frames) for 60 epochs on the VIS domain. Because the SSAE is pretrained once and transfers across datasets and spectral domains, and the heavy backbone is reused rather than retrained, the overall training cost remains modest despite the model's total size.

\begin{table}[tbp]
\centering
\caption{Two-stage training cost on a single RTX 5090.}
\label{tab:supp_training}
\begin{tabular}{lccc}
\toprule
Stage & Per epoch & Epochs & Supervision \\
\midrule
SSAE & 2.2~min & 150 & Self-supervised \\
Tracker & 13.9~min & 60 & Tracking labels \\
\bottomrule
\end{tabular}
\end{table}

\section{Failure Case Analysis}
\begin{figure}[tbp]
\centering
\includegraphics[width=\columnwidth]{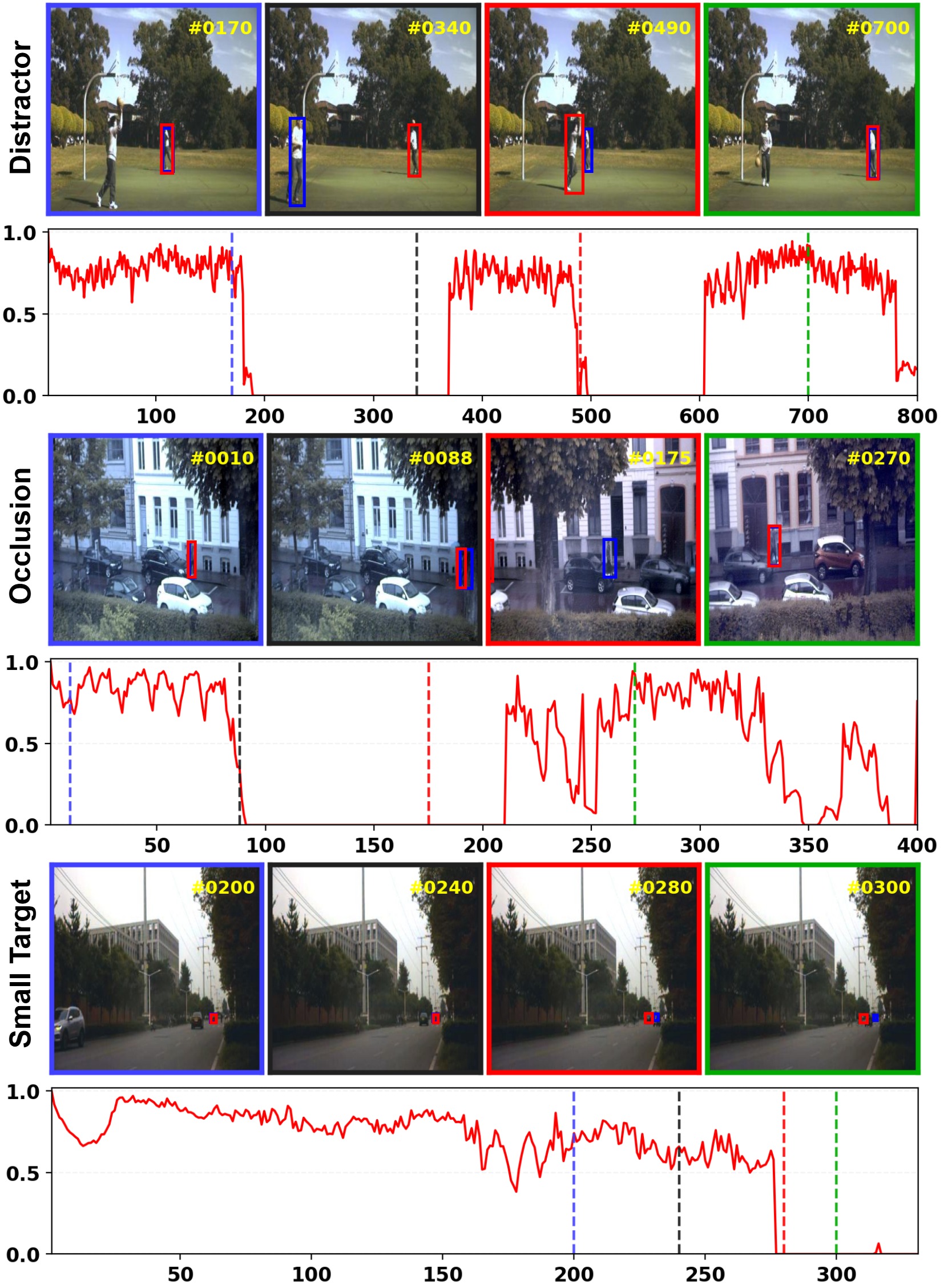}
\caption{Examples of failure cases and the IoU (overlap ratio) between predictions and ground truth. The colored dashed lines are corresponding to the frames above with borders in the same colors. Dashed blue boxes denote ground truth; red boxes are our predictions. Row 1 (Distractor): target-distractor interaction; Row 2 (Occlusion): prolonged full occlusion; Row 3 (Small Target): extremely small target.}
\label{fig:failure}
\end{figure}

While SProTrack achieves strong overall performance, we identify three systematic failure modes, illustrated in \cref{fig:failure}. These limitations are common among hyperspectral trackers that rely on visual backbones pretrained in the RGB domain.
 
\noindent\textbf{Target-distractor interaction.} When the target physically interacts with or passes closely behind a visually similar object, the predicted bounding box tends to shift toward the distractor, particularly when the distractor is larger or more salient. The score map produces a bimodal response during the interaction period, and the box regression head resolves this ambiguity by averaging the two peaks, resulting in a spatially offset prediction. Although the temporal token facilitates recovery after separation, the transient drift during interaction degrades frame-level metrics.
 
\noindent\textbf{Prolonged full occlusion.} Unlike the partial occlusion in \cref{tab:attribute}, here we refer to prolonged full occlusion. When the target is fully occluded for an extended duration (e.g., $>$20 consecutive frames), the temporal token gradually accumulates stale appearance information, causing the tracker to either freeze at the last known position or drift toward a nearby distractor with similar appearance. This limitation is inherent to the frozen ODTrack\cite{odtrack} backbone, which lacks an explicit re-detection mechanism.
 
\noindent\textbf{Extremely small targets.} For targets occupying fewer than approximately 100 pixels, both the false-color image and the SSAE latent representation lack sufficient spatial resolution to encode discriminative spectral features. In such cases, the spectral prompt contributes minimal additional information, and our method degrades to the performance level of the baseline ODTrack\cite{odtrack}.

\noindent\textbf{Analysis.} We stress that these failure modes are inherited from the frozen RGB backbone (lack of re-detection) or from input resolution limits (small targets), not from the spectral representation itself. In the appearance-variation and motion-dynamics regimes where spectral cues are most informative, deformation, occlusion onset, or illumination change, SProTrack ranks first (\cref{tab:attribute}), confirming that the spectral advantage holds precisely where it is expected to, while the residual failures lie outside the spectral module's scope.

\section{Statistical Reliability Analysis}

HOT2020 contains only 35 test sequences, and sequence difficulty varies far
more than method quality does. Marginal per-sequence standard deviations are
therefore dominated by a variance component shared across all trackers and
overlap heavily even between clearly different methods. We instead report a
paired analysis (Tab.~\ref{tab:paired}), taking DP differences \emph{within}
each sequence against hyperspectral baselines, with a percentile bootstrap 95\%
CI and a two-sided Wilcoxon signed-rank test. SProTrack attains a higher mean paired DP against every baseline and wins on
more sequences than it loses in all four comparisons; the comparison against
the strongest baseline, DaSSPNet, is significant ($p=0.012$). Power is limited
here for two reasons: DP@20 saturates at $1.0$ on easy sequences, leaving 12--15
ties per comparison, and one outlier: a small-target failure
of the kind costs $0.32$--$0.37$ DP against every
baseline. All intervals include zero while the signed-rank test rejects for
DaSSPNet; this is expected, since the interval is computed on the mean, which a
few large negative differences shift, whereas the signed-rank test reflects the
direction of the differences.

\begin{table}[t]
\centering
\caption{Paired per-sequence comparison on HOT2020. Differences in
  DP@20 are taken within each sequence, which cancels the difficulty term shared by all trackers. CI is a percentile bootstrap over sequences (10{,}000 resamples); $p$ is a two-sided Wilcoxon signed-rank test. Ties are frequent because DP@20 saturates at $1.0$ on easy sequences, so we report win/tie/loss (W/T/L) counts.}
\label{tab:paired}
\setlength{\tabcolsep}{4pt}
\small
\resizebox{\linewidth}{!}{\begin{tabular}{lcccc}
\toprule
Method & $\Delta$DP & 95\% CI & $p$ & W/T/L \\
\midrule
vs.\ MMF-Net~\cite{li2024materialmmf} & $+0.0212$ & $[-0.0137, +0.0546]$ & 0.114 & 14/12/9 \\
vs.\ SPIRIT~\cite{spirit} & $+0.0283$ & $[-0.0126, +0.0707]$ & 0.123 & 13/13/9 \\
vs.\ SEE-Net~\cite{seenet} & $+0.0199$ & $[-0.0163, +0.0572]$ & 0.108 & 14/15/6 \\
vs.\ DaSSPNet~\cite{dasspnet} & $+0.0357$ & $[-0.0049, +0.0907]$ & 0.012 & 17/13/5 \\
\bottomrule
\end{tabular}}
\end{table}

\section{Limitations}
\label{sec:limitation}

SProTrack has several limitations. First, its higher MACs and parameter count, inherited mainly from the frozen ODTrack\cite{odtrack}, may limit deployment on constrained hardware. Nevertheless, only 0.80\% of parameters are trainable, and template latent caching enables real-time inference (43.1 FPS), trading static cost for stronger temporal priors under data scarcity. Second, we train a separate SSAE for each spectral domain rather than jointly across VIS, NIR, and RedNIR; cross-domain adaptation, as explored by DaSSPNet\cite{dasspnet}, remains future work. Finally, the frozen backbone lacks explicit re-detection, causing failures under prolonged occlusion, while spectral prompting provides limited benefit for extremely small targets.

\section*{LLM Usage Declaration}
Large language models were used solely as writing assistance tools during the preparation of this manuscript. Specifically, they were used for grammar correction, sentence rephrasing, and improving the clarity of English expressions. 

All scientific contributions, including the research idea, method design, experimental setup, analysis, and interpretation of results, were conceived and conducted entirely by the human authors. No large language model was used to generate novel content, and none is considered an author or co-creator of this work.

\end{document}